# Unsupervised Text Mining of COVID-19 Records

Mohamad Zamini
dept.of Computer science
*University of North Dakota*
Grand Forks, ND, USA
mohamad.zamini@und.edu

*Abstract*— Since the beginning of coronavirus, the disease has spread worldwide and drastically changed many aspects of the human's lifestyle. Twitter as a powerful tool can help researchers measure public health in response to COVID-19. According to the high volume of data production on social networks, automated text mining approaches can help search, read and summarize helpful information. This paper preprocessed the existing medical dataset regarding COVID-19 named CORD-19[1] and annotated the dataset for supervised classification tasks. At this time of the COVID-19 pandemic, we made a preprocessed dataset for the research community. This may contribute towards finding new solutions for some social interventions that COVID-19 has made. The preprocessed version of the mentioned dataset is publicly available through Github[2].

## I. INTRODUCTION

World Health Organization (WHO) has announced COVID-19 as a pandemic regarding severity, fast-spreading, and inaction on March 11, 2020. Up to July 6, 2021, over 184 million cases were confirmed having COVID-19 and 4 million confirmed deaths caused by this virus. COVID-19 can cause severe respiratory syndrome and can be transmitted through droplets and direct contact. To fight the pandemic, preventive measures are executed by different countries to reduce its spread. The major steps like lockdowns, social distancing, stay-at-home norms, wearing masks can be mentioned. These actions resulted in negative impacts on human life like wide unemployment. Losing jobs spiked high levels of stress, and this can be observed from social networking websites like Twitter, Facebook, etc. which people try to describe their emotions there.

It is essential to measure public health and limit human-to-human interactions. During pandemics people tend to stay at home and surf social media more. Facebook and Twitter have become a more active source of information extraction among many social media platforms because of spreading breaking news faster than official news channels[1]. The amount of data generation can range from hundreds of thousands to millions which can extract useful information for an effective response to a crisis. With 500 million Tweets sent each day, Twitter is a source of information extraction relating to any crisis. Previous works regarding Padang Indonesia[2], tsunami hazards during the 2012 earthquake[3], and the 2013 Pakistan Earthquake[4] showed it could provide good insight into the event by designing machine learning models for classifying into various categories.

The information flow has encouraged many researchers to participate in the analysis of Tweets in various domains.

However, the process of data collection and processing from Twitter or other resources (e.g., PubMed) is time-consuming in addition to programming knowledge requirements. The motivation of this paper is to provide preprocess and publicly available datasets from the unstructured medical text plus data annotation. The methodology is applied on a ready-to-use COVID dataset for future data analysis tasks like classification or sentiment analysis. The dataset provided by the [5] from multiple resources and literature, including PubMed Central [5]. The dataset includes various information such as resource title, abstract, ID, authors, and journals [5]. Table 1. Provides the overview of the sample of the dataset regarding the COVID disease description.

Twitter data acquisition is the process of data collection from Twitter streams and providing a dataset in a meaningful format like comma-separated (.csv) and JavaScript Object Notation (.json). The Twitter Application Programming Interface (API) is a useful tool compatible with executing complex queries, including search and extract Tweets related to certain topics within a special period of time. To use Twitter API, you need to have an active Twitter user account with some restrictions, like Twitter contents are available for non-commercial research purposes. For protecting the privacy of users, and their personal information, just Tweet IDs are available in the form of a spreadsheet. These Tweet IDs are in a unique integer form, and for converting them to meaningful Tweets, a hydration process like SMMT tool or DocNow is needed. Both of the mentioned tools are open-source software models

In [6] gathered over 92 million Tweets from 8.9 million unique users with 5000 annotated Tweets for training the machine and quantified the public health beliefs. Understanding general public health beliefs expression can help the pandemic management improve. They trained various classifiers, including Ridge classifier, perceptron, passive-aggressive classifier; k-nearest neighbors classifier, random forest; support vector machine on the annotated data, and used AUROC as their metric.

In another work by [7], they analyzed 4 million Tweets related to COVID-19 using COVID-19 related hashtags like coronavirus, COVID-19, quarantine. They used Latent Dirichlet Allocation to find important unigrams (like a virus, lockdown, quarantine), bigrams (like social distancing, stay home, coronavirus), and sentiments.

## II. RELATED WORK

### A. Tweets dataset

Multiple studies shared datasets like [8], which contain 524 million Tweets, [9] has approximately 1.12 billion Tweets (about 287 million unique Tweets with no retweet), or [10] has published 123 million Tweets with over 60% English Tweets which are multilingual so for researchers.

---

[1] https://www.semanticscholar.org/cord19/download
[2] https://github.com/mzamini92/preprocessed-cord-19

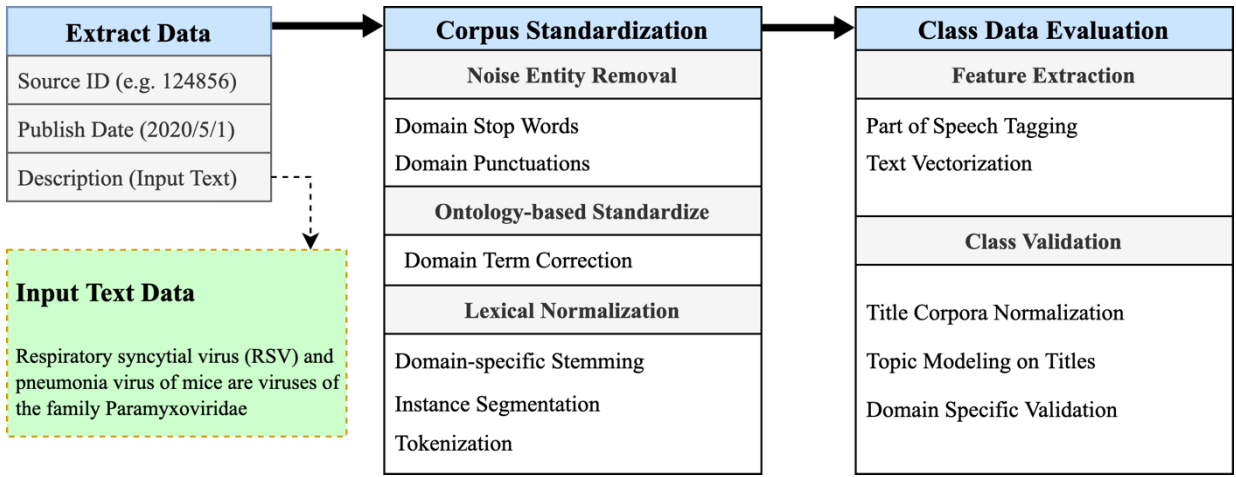

Fig. 1. Pipeline of preprocessing and evaluation of COVID dataset.

who need only English Tweets it may cause some difficulties in filtering. [11] provides 1.4 billion Tweet IDs in English, which needs to be fetched their Tweets using Twitter API. In response to the COVID-19 pandemic, Microsoft research, Allen Institute for AI, IBM and some other institutes prepared the CORD-19 dataset. It consists of scientific literature containing 141,000 articles related to Corona viruses.

*B. Sentiment analysis*

Some studies have recently been done on Twitter sentiment analysis like [12] which analyzed around 2.8 million COVID Tweets from nearly 160 thousand users from February 2020 to March 2020. They analyzed Tweets by using unigram and bigram frequencies and performed sentiment analysis. The results show the response of public health crisis in real-world and online are in the same way.

In [13] Used roughly 14 million Tweets for topic modeling in 26 topics and average compound sentiment scores were found to be harmful to the issues of spread and growth of case, symptoms, etc. and found sentiment from negative to positive for prevention, the impact of economy and markets, etc.

In [14] aimed to examine trends of four emotions, including fear, anger, sadness, and joy, on over 20 million Tweets from January 28, 2020, to April 9, 2020. The results show emotions changed from fear to anger through time and need to balance public psychological intervention.

*C. Social neTwork analysis*

Hashtags can help delegitimize shared content containing conspiracy theories. They need to be prevented and removed and support public health. In [15] they implemented social network analysis using the #5GCoronavirus hashtag to develop 5G conspiracy theories and strategies to deal with misinformation. They analyzed the nodes using the betweenness centrality score and grouped by Clauset-Newman-Moore algorithm clustering. The results showed a lack of authority to combat this misinformation.

In another work by [16], they analyzed online topics and discourse for the Protest movement in reaction to social distancing by content-coding Tweets and topic clustering using a machine learning algorithm on Twitter. Finally, the social network analysis that was used showed Protester Supporter network had a more centralized structure, but the Non-Supporter network has many small and exact size nodes.

In [17], they studied the role of public key players in social networks based on 2864 users, and it showed President Trump plays the most critical role in social networks in the top 20 key players.

The study of the #FilmYourHospital hashtag tried to understand key players and determine which online sources used more as evidence of their theory by using 22785 Tweets and 11333 users. The result showed Youtube is the main source of users.

*D. Topic modeling*

Topic modeling method applied in various works. In [18] they studied COVID Nonpharmaceutical interventions (NPI) in six countries and the trends in public attitude toward NPIs by using around 778 thousand English Tweets. Applying Pearson correlation analysis to extract relation between Tweet frequencies and case numbers and topic modeling to isolating Tweets about NPI showed New Zealand displayed the greatest attention to NPIs and US showed the least.

In another work by [19], they analyzed Twitter narratives around federal and state-level decision-making in response to COVID-19 is the US. They used 73 politicians' Tweets from January 1, 2020, to April 7, 2020, which obtained 7881 COVID-related Tweets and applied neTwork Hawkes binomial topic model to find evolving sub-topics about risk, testing, and treatment.

Dealing with the unstructured dataset is also investigated in non-medical domains. In [20], authors studied various preprocessing and topic modeling approaches to clean the aviation, automotive, and facility domains datasets. The datasets in a domain such as aviation and automotive contain many abbreviations and domain-specific vocabularies where require an appropriate preprocessing method to convert them to structured format. The authors developed a novel preprocessing pipeline for cleaning the non-medical domain-specific dataset where it motivated this work to deal with domain-specific dataset in the medical domain.

In [21], authors studied topics and sentiments about COVID-19 vaccine discussions by using approximately 1.5 million unique Tweets from 583,499 users. They used Latent

Dirichlet Allocation for topic modeling, and the result showed 16 topics grouped into five overarching themes.

## III. DATASET INTRODUCTION

The collected data are from PubMed Central (PMC), PubMed, World Health Organization's COVID-19 Database, and preprint servers bioRxiv, medRxiv, arXiv. All provided papers are extracted with this query:

"COVID" OR "COVID-19" OR "Coronavirus" OR "Corona virus" OR "2019-nCoV" OR "SARS-CoV" OR "MERS-CoV" OR "Severe Acute Respiratory Syndrome" OR "Middle East Respiratory Syndrome" [5].

TABLE 1: ORIGINAL DESCRIPTION OF THE DISEASE PROVIDED IN THE SAMPLE DATA

| ID | Description | Date |
| --- | --- | --- |
| 111 | Latest updates on **COVID**-19 from the European Centre for Disease Prevention and Control | 2020-05-01 |
| 112 | Updated rapid risk assessment from **ECDC** on the outbreak of **COVID**-19: increased transmission globally | 2020-07-06 |
| 113 | Effects of respiratory rate on venous-to-arterial **CO(2)** tension difference in septic shock patients undergoing volume mechanical ventilation | 2020-08-12 |
| 114 | The novel coronavirus (**SARS-CoV**-2) infections in China: prevention, control and challenges | 2020-08-10 |
| 115 | Comparison of the structural protein coding sequences of the **VR**-2332 and Lelystad virus strains of the **PRRS** virus | 2020-09-02 |

## IV. METHODS AND MODELS

Dealing with the unstructured format of the Covid-19 text dataset is a key issue when developing models that use off-the-shelf NLP tools. Using various abbreviations and domain-specific terminologies is not specific to the biomedical domain text data, where authors in [20, 22, 23], developed the state-of-the-art preprocessing and annotation pipeline called "MaintNet" [22] that helps to standardize the domain-specific data and cluster them into semantically similar groups. In this work, to standardize the Covid dataset, MaintNet pipeline was employed, and the following sections provide the detail of the processing:

### A. Data Standardization

Regarding the pipeline, as shown in Figure 1, initially, the text normalization technique, including text lowercasing, was applied, followed by removing the domain-specific stop words and punctuation [20, 22, 23]. In order to expand the abbreviations, the domain-specific dictionary was manually curated by annotating the domain keywords and expanded based on the medical terms. The existing domain-specific abbreviations are expanded using scientific papers in this area and other online medical resources.

### B. Preprocessing

Furthermore, the normalized dataset was lemmatized using WordNet lemmatizer and tokenized with NLTK library as recommended by MaintNet [23]. Further, the term frequency-inverse document frequency (TFIDF) method was applied to the tokenized data to extract the essential features for the text clustering phase. For further evaluation of the titles that briefly describe the dataset's abstract content, text normalization and topic modeling can be applied for supervised classification tasks.

## V. RESULTS

As shown in Table 2, the raw and preprocessed version of the dataset representing the written abstract description provided. The outcome of the preprocessing confirms the outperformance of the MaintNet [22] preprocessing pipeline in converting the domain-specific abbreviations and terms to the proper standard format.

TABLE 2: ORIGINAL AND PREPROCESSED SAMPLE DATA

| Raw version | Preprocessed version | Date |
| --- | --- | --- |
| coronavirus disease (covid) presents two urgent health problems: the illness caused by the virus itself and the anxiety, panic and psychological problems associated with the pandemic. | coronavirus disease coronavirus disease present two urgent health problems illness cause by virus and anxiety panic and psychological problems associate with pandemic | 2020-04-15 |

Furthermore, Figure 2 presents the Word-Cloud of the preprocessed sample of the 100,000 datasets showing the main domain terms that are mostly utilized in this dataset. The outcome of this preprocessed dataset and domain-specific abbreviations and terms were made available to the public Github.

In the future direction of this research, supervised classification tasks such as entity recognition and authorship classification will be investigated. The methodologies and algorithms such as convolutional neural networks developed by recent work [24] will be examined into these domain-specific medical datasets.

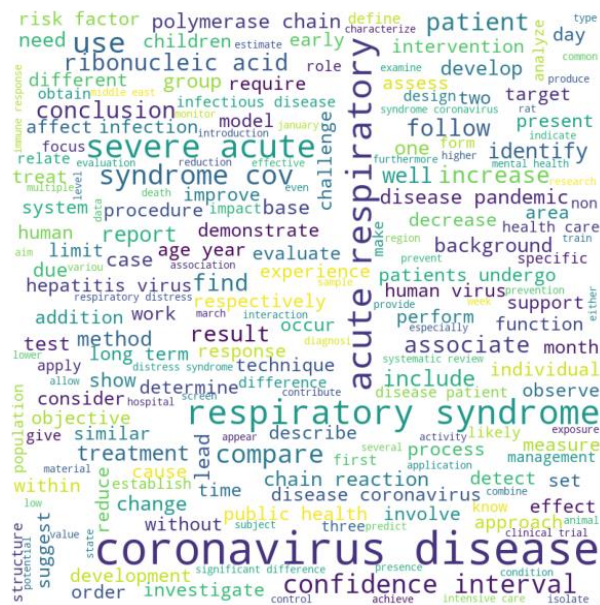

Fig. 2. Visualization of the word cloud representing the various domain specific terms in medical domain.